\documentclass[letterpaper, 10 pt, conference]{ieeeconf}  

\IEEEoverridecommandlockouts                              

\usepackage{graphicx}
\usepackage{subfig}
\usepackage{float}
\usepackage{amsmath} 
\usepackage{amssymb}  
\usepackage{balance}
\usepackage{bm}
\usepackage{multirow}
\usepackage{comment}
\usepackage{url}

\newcommand{\vect}[1]{\boldsymbol{#1}}
\DeclareMathOperator*{\argmax}{argmax}
\newcommand*{\permcomb}[4][0mu]{{{}^{#3}\mkern#1#2_{#4}}}
\newcommand*{\perm}[1][-3mu]{\permcomb[#1]{P}}

\title{\LARGE \bf
Multimodal Coherent Explanation Generation of Robot Failures
}

\author{Pradip Pramanick$^{1}$ and Silvia Rossi$^{1,2}$
\thanks{This work has been partially supported by the European Union's Horizon Europe research and innovation programme under the TRAIL project, Marie Sk\l{}odowska-Curie grant agreement No 101072488, and by the Italian Ministry for Universities and Research (MUR) with the PNRR Project FAIR (Future Artificial Intelligence Research) PE0000013.}
\thanks{$^{1}$Pradip Pramanick is with the Interdepartmental Center for Advances in Robotic Surgery - ICAROS, University of Naples Federico II, Naples, Italy {\tt\small pradip.pramanick@unina.it}}%
\thanks{$^{2}$Silvia Rossi is with the Department of Electrical Engineering and Information Technologies - DIETI, University of Naples Federico II, Napoli, Italy {\tt\small  silvia.rossi@unina.it}}%
}

\begin{document}

\maketitle
\thispagestyle{empty}
\pagestyle{empty}

\begin{abstract}
The explainability of a robot’s actions is crucial to its acceptance in social spaces. Explaining why a robot fails to complete a given task is particularly important for non-expert users to be aware of the robot's capabilities and limitations. So far, research on explaining robot failures has only considered generating textual explanations, even though several studies have shown the benefits of multimodal ones. However, a simple combination of multiple modalities may lead to semantic incoherence between the information across different modalities - a problem that is not well-studied. An incoherent multimodal explanation can be difficult to understand, and it may even become inconsistent with what the robot and the human observe and how they perform reasoning with the observations. Such inconsistencies may lead to wrong conclusions about the robot's capabilities. In this paper, we introduce an approach to generate coherent multimodal explanations by checking the logical coherence of explanations from different modalities, followed by refinements as required. We propose a classification approach for coherence assessment, where we evaluate if an explanation logically follows another. Our experiments suggest that fine-tuning a neural network that was pre-trained to recognize textual entailment, performs well for coherence assessment of multimodal explanations. Code \& data: \url{https://pradippramanick.github.io/coherent-explain/}.

\end{abstract}

\section{Introduction}
With the growing potential of assistive robotics, there is an increasing concern about the explainability of the decisions they make and the predictability of the outcome of such decisions. These concerns are amplified when the robot's behavior is decided by complex systems that are often non-deterministic and with the possibility of failures in unexpected situations. For non-expert users and observers, understanding the reason for a failure can set realistic expectations for the robot and help to build trust~\cite{nesset2021transparency}.

The existing works on providing explanations for robot failures primarily use text as a modality~\cite{das_explainable_2021,liu2023reflect}. While explaining using natural language can be intuitive to a non-expert user, there are several limitations in using only text as an explanation medium~\cite{wallkotter_explainable_2021}. Prior studies reveal a need for multi-modal explanation~\cite{anjomshoae2019explainable} and highlight its benefits in terms of intuitiveness and efficiency in presenting complex information~\cite{park_multimodal_2018,chen_rex_2022,zhu_affective_2022}. However, the problem of multi-modal explanation is not well-studied in the context of providing explanations of robot failures. Further, previous research on multimodal explanation generation does not study the coherence of the generated explanations across the modalities, even though incoherent explanations can occur in several scenarios, as we discuss later.

In this paper, we present the problem of coherent multimodal explanation generation of robot failures. Particularly, we study a combination of two modalities:
\begin{enumerate}
    \item A text modality that contains a natural language description of an observed failure;
    \item A graphic modality that shows information about the cause of the failure, such as the robot's plan until the failure and beliefs represented as a scene graph, overlaid on the robot's egocentric-view image that captures failure observation.
\end{enumerate}

\begin{figure*}[t]
    \centering
    \subfloat[Textual explanation generated by~\cite{liu2023reflect}.]{\includegraphics[width=0.75\linewidth]{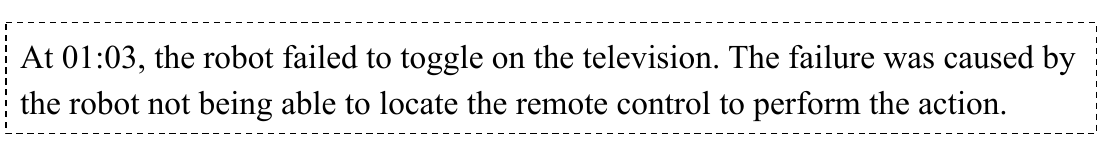}\label{fig:1text}} \\
    \subfloat[Graphical explanation is contradictory to textual explanation.]{\includegraphics[width=0.45\linewidth]{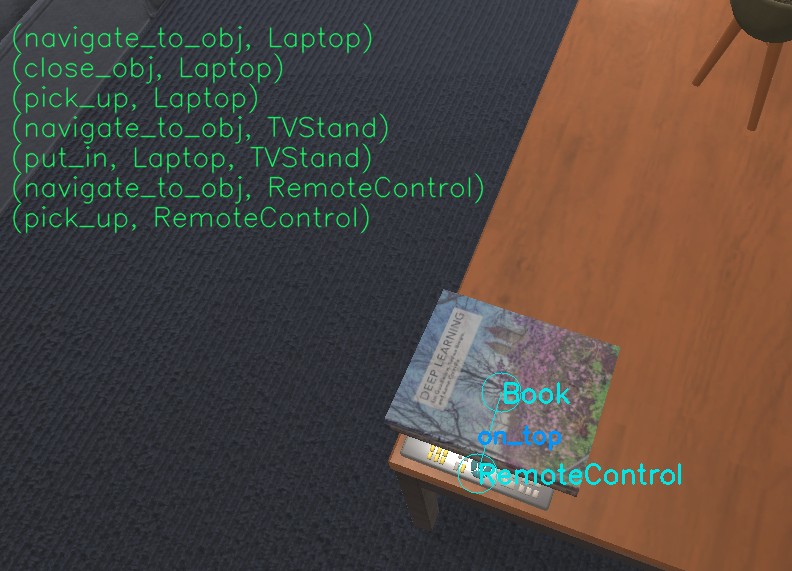}\label{fig:1contradict}}
    \qquad \qquad
    \subfloat[Observation at 01:03 does not reflect the cause of the failure.]{\includegraphics[width=0.45\linewidth]{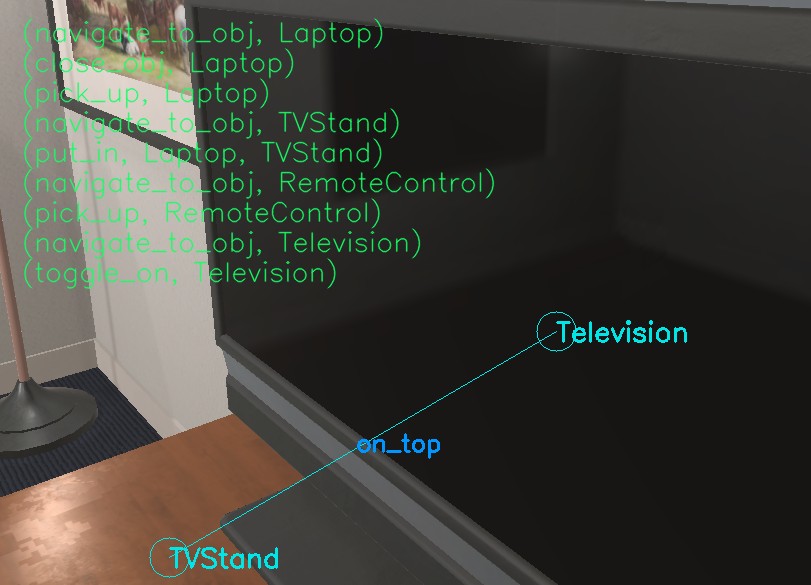}\label{fig:1incoherent}}
    \caption{Lack of coherence between textual and graphical modalities in explanation of robot failure.}
    \label{fig:motivation1}
\end{figure*}

Our initial observation indicates that a simplistic amalgamation of modalities may result in scenarios where the information presented across the two modalities is inconsistent. There are two major reasons for this. Firstly, such inconsistencies may stem from text generation using neural networks, where even large language models (LLMs) tend to hallucinate~\cite{liu2022token}, despite showing strong reasoning skills at times for explanation generation~\cite{liu2023reflect}. As a motivating example, consider Figure~\ref{fig:1text} which shows a textual explanation generated by a state-of-the-art method~\cite{liu2023reflect} for failing to turn on a television. The textual explanation incorrectly says the robot could not locate the remote control, which contradicts the robot's belief (i.e., the robot detects the remote control below the book) and its visualization in the graphic modality, as shown in Figure~\ref{fig:1contradict}. Second, the world model of the robot and the explanation generator (e.g., a LLM) can be different. This leads to situations where the reasoning in the text explanation is based on a human's approximation\footnote{This is because the LLM is trained on human-generated data.} of a robot's world model, while valid in a human's world model, may not apply to a specific robot having a specific set of available actions, or different observation capabilities. This phenomenon is similar to the problem of \textit{Model Reconciliation} in the explainable planning literature and has been extensively studied using formal methods \cite{sreedharan2021foundations}.

Another reason for a lack of coherence between the two modalities is the possibility of an observed failure and the cause of the failure not coinciding temporally. We show another motivating example in Fig.~\ref{fig:1incoherent}, where for the same task, the failure occurs at 01:03, but the robot's observation at the time does not reflect the cause of the failure. Such incoherent explanations may induce incorrect beliefs about the robot and its capabilities, which may further affect the ability to trust the robot's explanations. 

In this paper, we present a method to detect such inconsistencies and refine the multimodal explanation to make it coherent. Our main contributions are summarized in the following.
\begin{itemize}
    \item Conceptually, we formulate the problem of coherence assessment in multimodal explanations, which requires reasoning with multimodal information. 
    \item Technically, we propose an approach to the evaluation of coherence in multimodal explanations of robot failures as a classification task, and we discuss strategies to refine incoherent explanations.
    \item Empirically, we find that transfer learning on a related problem of textual entailment recognition, combined with counterfactual training examples, leads to efficient training of a neural reasoner that can accurately detect coherence in multimodal explanations.   
\end{itemize}

\section{Related Work}
\label{sec:related}
A large body of prior research on explainable artificial intelligence (XAI) focuses on improving the transparency of black-box classifiers, i.e., they provide methods for reasoning over a single instance of a decision-making problem~\cite{koh2020concept}, or even a set of non-sequential instances~\cite{setzu2021glocalx}. In contrast, the explainability of robot behavior generally involves explaining a sequential decision-making problem. Recent reviews on explainable robotics~\cite{anjomshoae2019explainable,wallkotter_explainable_2021} provide a summary of methods, application areas, and evaluation methodologies for explaining robot behavior. They also highlight the lack of research on the explanation of robot failures and multimodal approaches to explanation. The following summarizes relevant research on these two topics, along with research on multimodal coherence.

\subsection{Failure Explanation}
Autonomous failure detection is often a precursor to explanation. Several approaches have been proposed to do so, which include both model-based reasoning~\cite{knepper2015recovering,chen2020rprs} and data-driven learning to predict anomalies that often take multimodal sensory data as input~\cite{altan2021went,inceoglu2021fino}. While these approaches are important contributions to detecting both planning and execution failures, they are limited to failure detection without explanation. Similarly, research on the explanation of failures can be broadly categorized into three approaches, which generally involve finding a cause for an observed failure. 

First, model-based approaches perform reasoning with a formal world model and symbolic observations~\cite{nair2020feature}, and focus on providing contrastive explanations for planning~\cite{sreedharan2021foundations} and sub-optimal behavior~\cite{tulli2020explainable}, not considering execution failures. Also, the recipients of the explanations are domain experts, instead of non-expert users that we target in this work. Second, data-driven methods learn from labels provided by non-experts to automatically generate explanations from a sequence of prepossessed sensory observations. Inceoglu et al. model explanation of manipulation failures as a failure-type classification problem~\cite{inceoglu2023multimodal}. Further,~\cite{das_explainable_2021,ehsan2019automated} propose methods for learning to generate textual explanations of failures. Finally, neural-symbolic approaches encode domain knowledge using symbolic constructs to either formulate a data-efficient learning problem, for both experts~\cite{diehl2022did} and non-experts~\cite{liu2023reflect,gavriilidis2023surrogate}, or convert state predictions into explanations using templates~\cite{chiyah2021self,das2021semantic}. However, most of the previous approaches to generating explanations of robot failures only consider a single modality, i.e., text. We consider a recently published work~\cite{liu2023reflect} as the state of the art in textual explanation generation of failures for our experiments.
\subsection{Multimodal Explanation}
Prior studies in HRI suggest that multimodal explanations are often more efficient and intuitive than unimodal explanations, particularly compared to textual explanations~\cite{wallkotter_explainable_2021,alipour2020study,schott_literature_2023}. Several works have addressed the problem of explaining the answers to visual question-answering (VQA) systems, by providing visual evidence along with textual explanations~\cite{park_multimodal_2018,chen_rex_2022}. VQA explanations are relevant for the problem addressed in this paper since they perform reasoning over a sequence of predictions. In robotics, several works have explored the combination of text and some form of graphics to improve the transparency and explainability of robotic systems and classifiers used for HRI. Perlmutter et al.~\cite{perlmutter_situated_2016} combined visualization of a robot's beliefs and intentions with textual feedback to improve the transparency of a situated language understanding system. A similar form of visualization has been explored in~\cite{zhu_affective_2022} to explain emotion recognition in HRI and in~\cite{wang2022investigating} for an explainable HRI system to teach robots with augmented reality. Hastie et al. developed a multimodal interface by combining text explanations in a graphical interface for transparent interaction with a remote robot~\cite{hastie2018miriam}. Even though the prior research on multimodal explanations has not been specifically applied to failure explanations, our selection of modality combinations for studying coherence is motivated by these.

\subsection{Multimodal Coherence}
The majority of research on computational models of multimodal coherence focuses on image-text coherence. Several taxonomies have been proposed, primarily based on the theory of discourse relations in linguistics~\cite{prasad2008penn}. 
Otto et al. propose a categorization of semantic relations between images and text and a method to detect them~\cite{otto2019understanding}. This categorization is based on three attributes — cross-modal mutual information, the presence of hierarchy, and semantic correlation, which is analogous to our definition of coherence. Alikhani et al. propose six classes of coherence relations based on an image captioning dataset~\cite{alikhani2020cross}. They also present a method for predicting the relations and a coherence-aware image captioning model. These relations are further analyzed in~\cite{alikhani2023image}, along with an evaluation of several vision-language models for the task of predicting the relations. The taxonomy in~\cite{alikhani2020cross} and~\cite{alikhani2023image} is almost comprehensive for textual descriptions of images, except it does not consider contradiction. 

Further, much of the existing taxonomies are not formally defined, leading to subjective interpretation and classification ambiguity, e.g., multiple relations are applicable for the same pair of image and text descriptions~\cite{alikhani2020cross}. In contrast, our model of coherence assessment focuses on semantics, instead of expressiveness or the style of description. Thus, it is simpler, less ambiguous, and allows us to model coherence assessment as an entailment recognition problem. In this regard, our work is also relevant to multi-modal stance detection~\cite{yuan2023support} and fact-checking~\cite{yao_end--end_2023}, which follow a similar taxonomy. However, the existing taxonomies and methods for their prediction are designed for problems such as image captioning and multimodal information retrieval, and thus they cannot be trivially applied to the problem of coherence in multimodal explanation of robot failures.   

\section{Proposed Framework}
\label{sec:model}
In this section, we first formally introduce the multimodal explanation framework and define the problem, before describing our methodology in detail. Given a high-level task plan $\pi$, a sequence of observations taken at $n$ discrete time steps $\bm{O} = \{ O_1, O_2, \dots, O_n\}$, where a failure is observed in $O_i$, we aim to present a multimodal explanation $\mathcal{E}^m$ for $\pi$. $\mathcal{E}^m$ consists of a pair of mutually coherent explanations, a textual explanation $\mathcal{E}^t$ and a graphical explanation $\mathcal{E}^g = \{ \mathcal{E}^\pi \cup \mathcal{E}^{O_i} \}$, which is overlaid on the corresponding ego-view image $I_i$ at time step $i$. Each $\mathcal{E}^g$ has two components, an explanation of action execution, i.e., the plan until the failure observation $\mathcal{E}^\pi$, and a sub-graph of the scene graph at $i$, $\mathcal{E}^{O_i}$. To obtain $\mathcal{E}^m$, we first obtain a base $\mathcal{E}^t$ and $\mathcal{E}^g$ independently, by reasoning over $\pi$ and $\bm{O}$, which we describe in Section~\ref{sec:base-explanation}. Next, we assess the coherence between $\mathcal{E}^t$ and $\mathcal{E}^g$ and then perform refinements to either $\mathcal{E}^t$ or $\mathcal{E}^g$, as and if required. In the following, we formally define the two sub-problems.

\subsection{Coherence Assessment} Given a base textual explanation $\mathcal{E}^t$, a base graphical explanation at step $i$, $\mathcal{E}^g$, and the observation sequence $\bm{O}$, we model coherence assessment as a ternary classification task from the set,
\[ C= \{\mathcal{E}^g \models \mathcal{E}^t, \mathcal{E}^g \not \models \mathcal{E}^t, \mathcal{E}^g \bot \mathcal{E}^t\}, \]

where $\mathcal{E}^g \models \mathcal{E}^t$ denotes that $\mathcal{E}^t$ is supported by $\mathcal{E}^g$ and thus $\mathcal{E}^m = \{ \mathcal{E}^t \cup \mathcal{E}^g\}$ is coherent, $\mathcal{E}^g \not \models \mathcal{E}^t$ denotes that $\mathcal{E}^t$ is not supported by $\mathcal{E}^g$, and $\mathcal{E}^g \bot \mathcal{E}^t$ denotes that $\mathcal{E}^t$ contradicts $\mathcal{E}^g$. Particularly, we want to estimate the following,
\[ c = \argmax_{c \in C} P(c | \mathcal{E}^t, \mathcal{E}^g, \pi ) .\]
We describe the method to learn this classification in Section~\ref{sec:MCC}.

\subsection{Explanation Refinement}
Based on the outcome of the above classification, we either present the multimodal explanation as is, i.e., in the case of $\mathcal{E}^g \models \mathcal{E}^t$; or we select one of the following refinement strategies.
\begin{itemize}
    \item Refine — $\mathcal{E}^g \not \models \mathcal{E}^t$: This refinement strategy assumes that $\mathcal{E}^t$ is correct, and therefore searches for a new graphical explanation in a time step $j$, $\mathcal{E}^{g'}, \forall j \in \{0..n\} \setminus \{i\}$ that satisfies $\mathcal{E}^{g'} \models \mathcal{E}^t$.
    \item Refine — $\mathcal{E}^g \bot \mathcal{E}^t$:  This refinement strategy assumes that $\mathcal{E}^t$ is incorrect and therefore proposes a refined textual explanation $\mathcal{E}^{t'}$ that satisfies $\mathcal{E}^g \models \mathcal{E}^{t'}$.
\end{itemize}
We detail the refinement strategies in Section~\ref{sec:refinement}.
\subsection{Obtaining $\mathcal{E}^t$ and $\mathcal{E}^g$}
\label{sec:base-explanation}
We rely on the work of Liu et al.~\cite{liu2023reflect} to generate $\mathcal{E}^t$. More specifically, we convert the tuple ($\pi,\bm{O}$) into a natural language description using the method proposed in~\cite{liu2023reflect}. The natural language description of the plan and the observations is a sequence of tuples that consists of an action from the plan and the robot's observation after attempting to execute the action. We put this summary of action execution in a template and prompt a large language model\footnote{https://platform.openai.com/docs/models/gpt-3-5-turbo}, which generates a textual explanation of the failure, along with a prediction of the time step $i$. We use the same prompt templates as~\cite{liu2023reflect}. For completeness, we also perform experiments with the expert-provided failure time steps and explanations in~\cite{liu2023reflect}.

We generate $\mathcal{E}^{O_i}$ from $O_i$ and $\pi$. Specifically, we represent $O_i$ as a 2D scene graph using \cite{liu2023reflect}, and then perform a filtering operation to obtain a sub-graph,
\[ \mathcal{E}^{O_i} = v \in \mathcal{A}(\pi,i) \cup \{ v* \in V(O_i) : (v,v*) \in E(O_i)\}   \]
where $V(O_i)$ and $E(O_i)$ denote the set of vertices and edges in $O_i$ and the function $ \mathcal{A} (\pi,i)$ returns the arguments in the plan step (action) executed during $i$. This filtering returns a sub-graph where the vertices are either an argument of the action at $i$ or they have an edge with at least one of such vertices. We do this filtering to restrict the visualization of the scene graph to only the objects that are relevant to the current action, in an effort to highlight the cause of the failure concisely. We obtain $\mathcal{E}^{\pi}$ by simply selecting a sub-sequence of $\pi$ till $i$.

\subsection{Modality Coherence Classification}
\label{sec:MCC}
Before describing our method to perform coherence classification, we first define the class symbols in the context of our problem. Let us consider that an explanation $\mathcal{E}$ is a set of $m$ propositions, represented as a conjunction of grounded predicates,
\[ \mathcal{E} \equiv \mathcal{P}_0 \land \mathcal{P}_1 \land \dots \mathcal{P}_m .\]
Therefore, we define the class symbols as the following.
\begin{align*}
& \mathcal{E}^g \bot \mathcal{E}^t \equiv \exists \mathcal{P}_j \in \mathcal{E}^g, \exists \mathcal{P}_k \in \mathcal{E}^t :  \mathcal{P}_j \bot \mathcal{P}_k .\\
& \mathcal{E}^g \models \mathcal{E}^t \equiv \exists \mathcal{P}_j \in \mathcal{E}^g, \exists \mathcal{P}_k \in \mathcal{E}^t :  \mathcal{P}_j \models \mathcal{P}_k \land \neg  ( \mathcal{E}^g \bot \mathcal{E}^t ) .\\
& \mathcal{E}^g \not \models \mathcal{E}^t \equiv \neg ( \mathcal{E}^g \bot \mathcal{E}^t \land \mathcal{E}^g \models \mathcal{E}^t ).
\end{align*}
As an example, consider the following propositions from the observation in Figure~\ref{fig:1contradict}, \textit{on\_top(remote-control, table)} $\land$ \textit{on\_top(book, remote-control).} The proposition \textit{$\neg$(locate (remote-control))} in the text span “the robot not being able to locate the remote control'' is contradictory to \textit{on\_top (remote-control, table)}. Similarly, the proposition \textit{on\_top(book, remote-control)} entails the proposition \textit{is\_blocking(book, remote-control)} in the expert-written explanation “book is blocking the remote control” in ~\cite{liu2023reflect}. Finally, considering the propositions in the observation in Figure~\ref{fig:1incoherent} are \textit{on\_top(television, tv-stand)} $\land$ \textit{has\_state(television, off)}, the propositions in the text explanations are neither entailing nor contradicting.

\begin{figure}
    \centering
    \includegraphics[width=0.9\linewidth]{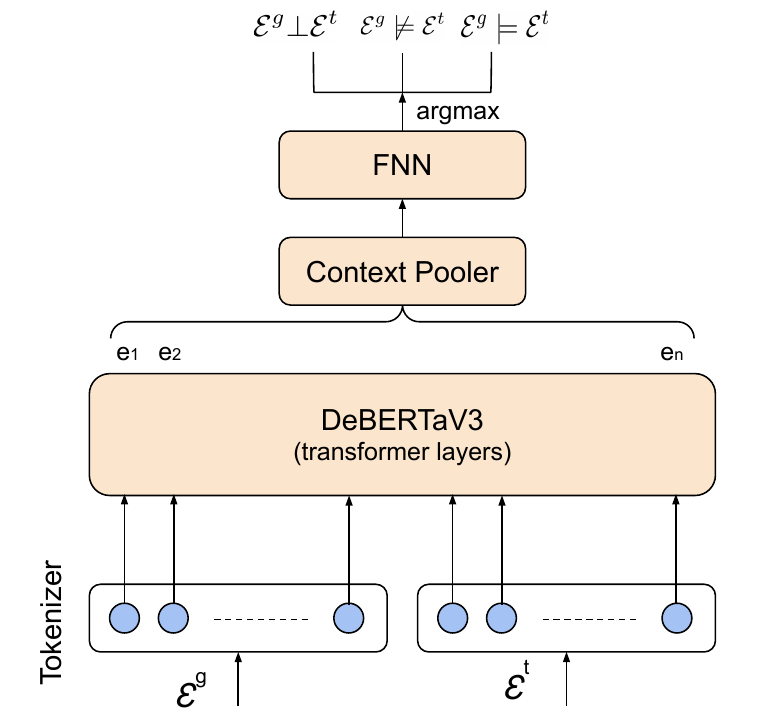}
    \caption{Our approach for learning coherence classification.}
    \label{fig:model}
\end{figure}

However, applying these rules requires prior knowledge of what propositions (or conjunction of propositions) are contradictory and entailing. Thus, we propose an approach to learn a neural reasoner, by training on annotations based on these rules. Our model, as shown in Fig.~\ref{fig:model}, takes a pair of inputs $\mathcal{E}^g, \mathcal{E}^t$, where $\mathcal{E}^g$ is one of $\mathcal{E}^\pi$ and $\mathcal{E}^{O_i}$. The tokenized inputs are passed through several transformer layers to obtain hidden representations $e_1,e_2,\dots,e_n$. We use a pre-trained DeBERTaV3~\cite{he2022debertav3} to obtain the hidden representations, which are further passed through a context pooler layer and finally through a feed-forward layer that learns to classify. 

We train the model jointly using annotated $\mathcal{E}^{O_i}, \mathcal{E}^t$ and $\mathcal{E}^\pi, \mathcal{E}^t$ pairs. For $\mathcal{E}^{O_i}$, the conjunction of propositions is a set, but for $\mathcal{E}^\pi$, it is a sequence. As our reasoning approach is somewhat similar to a well-studied problem in natural language processing, namely \textit{natural language inference} (NLI)~\cite{williams2018broad}, also known as recognizing textual entailment, we perform fine-tuning on a model that was pre-trained on multiple NLI datasets. Thus, we utilize the findings in the NLI domain to propose a data-efficient learning method for this reasoner. We provide details on this training in Section~\ref{sec:experiments}.

We can extend this reasoning from a pair of explanations to an arbitrary set of $l$ explanations $\vect{\mathcal{E}}$ by recursive applying the rules on explanations, instead of propositions. The method involves iteratively selecting an explanation $\mathcal{E}^x$ as the hypothesis, to compare with a conjunction of premises consisting of a subset of $\vect{\mathcal{E}}$, having $l-1$ explanations, $\{\mathcal{E}^Y : \vect{\mathcal{E}} - \mathcal{E}^x \}$. Thus, we can rewrite the reasoning rules as the following.
\begin{align*}
  & \mathcal{E}^Y \bot \mathcal{E}^x \equiv \exists \mathcal{E}^y \in \mathcal{E}^Y :  \mathcal{E}^y \bot \mathcal{E}^x .\\
  & \mathcal{E}^Y \models \mathcal{E}^x \equiv \exists \mathcal{E}^y \in \mathcal{E}^Y:  \mathcal{E}^y \models \mathcal{E}^x \land \neg  ( \mathcal{E}^Y \bot \mathcal{E}^x ) .\\
  & \mathcal{E}^Y \not \models \mathcal{E}^x \equiv \neg ( \mathcal{E}^Y \bot \mathcal{E}^x \land \mathcal{E}^Y \models \mathcal{E}^x ).
\end{align*}

Thus, by classifying both $\mathcal{E}^\pi, \mathcal{E}^t$ and $\mathcal{E}^{O_i}, \mathcal{E}^t$, we can apply the above rules to classify the pair $\mathcal{E}^g, \mathcal{E}^t$. A coherence assessment of a set of $l$ explanations requires $\perm{l}{2}$ comparisons. However, we can perform this reasoning more efficiently by early stopping and using other heuristics, such as assuming certain combinations are known or likely to be contradictory based on prior knowledge. However, experiments with more modalities are beyond the scope of this paper. 

\subsection{Refinement Strategies}
\label{sec:refinement}

\subsubsection{Refine — $\mathcal{E}^g \not \models \mathcal{E}^t$}
In this refinement strategy, we iteratively select a graphical explanation from the discrete time steps and perform a coherence assessment until we find a $\mathcal{E}^g$, such that $\mathcal{E}^g \models \mathcal{E}^t$ is satisfied. If we do not find such a time step, we fall back to finding a
$\mathcal{E}^g$ where either $\mathcal{E}^\pi \models \mathcal{E}^t$ or $\mathcal{E}^{O_i} \models \mathcal{E}^t$ is true. To do this efficiently, we restrict the selection of time steps to only those where a scene graph in $O_i$ is different from $O_{i-1}$. This is similar to the key-frame selection method in~\cite{liu2023reflect}. 
\subsubsection{Refine — $\mathcal{E}^g \bot \mathcal{E}^t$}
We propose a simple refinement strategy to generate the refined textual explanation for both $\mathcal{E}^\pi \bot \mathcal{E}^t$, and $\mathcal{E}^{O_i} \bot \mathcal{E}^t$. We refine the textual descriptions to a much simpler explanation of the task failure by the template - \textit{The robot failed to complete [TASK] because it was unable to perform [ACTION] at [TIME]}, where the [*] slots are filled by information for a particular task failure. This template is similar to several expert-written explanations in~\cite{liu2023reflect}. Even though this explanation template is less specific, we chose it to avoid providing contradictory explanations. We discuss a few ways to improve the refinement of textual explanations in Section~\ref{sec:discuss}.

\section{Experiments}
\label{sec:experiments}
\subsection{Data}
To evaluate our methods, we first obtain explanations from the \textit{RoboFail} dataset in~\cite{liu2023reflect}. We further generate counterfactual examples based on the metadata provided by the \textit{ai2thor} simulator~\cite{kolve2017ai2}. In the following, we describe these in detail.

\textbf{RoboFail Dataset (RF).}
RF contains various failure scenarios that are generated by manually injecting failure conditions in ai2thor. To utilize this dataset for evaluating our coherence classification model, we first extract tuples of the robot observation on the marked timestamp of failure, the plan until the failure, and the text explanations generated by~\cite{liu2023reflect}, as well as expert-written explanations. Then, we manually annotate the data using the definitions presented in Section~\ref{sec:MCC}, obtaining a total of 260 examples. We exclude explanations where the LLM fails to predict the time step of failure. To convert natural language text into a conjunction of propositions, we apply a heuristic method of converting the text into a predicate-argument structure using a pre-trained semantic parsing model~\cite{shi2019simple}.

\textbf{Counterfactual Generation (CF).}
RoboFail has a total of 29 examples of contradiction ($\approx$ 11\%). This is a significant percentage considering that the dataset was not developed to study coherence, it includes expert-written explanations, and even a few contradictions can negatively affect the explainability of a multimodal system. However, for training and a fair evaluation, we generate a larger and more balanced dataset by generating counterfactual examples. To do so,  we first sample a random task plan and a scene graph from RF. For sampling task plans, we restrict to this subset of RF tasks - \textit{boil water, heat potato, make coffee,} and \textit{toast bread}. Then we select a random failure type to inject from a subset\footnote{We select subsets to evaluate on unseen task and failure types.} of failure injection methods in RF - \textit{unexpected dynamics, failed execution, wrong order of actions} and \textit{missing actions}~\cite{liu2023reflect}. Next, depending on the failure type, we generate counterfactual examples by modifying either the plan, a set of observations, or both. More specifically, we perform modifications to the observation by replacing predicates, arguments, and adding negations. We further modify plans by introducing actions with unmet preconditions, by either deleting actions having a common effect, or by reversing a pair of actions having the same arguments. We collect a total of 1240 automatically annotated examples with counterfactual data generation.

To make training, validation and test sets, we separate RF into two subsets based on task types. We do this to test the generalizability of the reasoner on explanation pairs from unseen tasks. However, please note that even for the same task types, the propositions, or the conjunction of propositions are distinct. Additionally, for $\mathcal{E}^\pi$, the sequence of propositions is also distinct. Thus, we first separate the data of \textit{make salad, warm water,} and \textit{store egg} tasks from RF. There are 80 such data points which are absent in CF and not used in training and validation sets. The rest of the data in RF contains the task types \textit{water plant, cook egg,} and \textit{switch devices}, in addition to the four task types in CF. We combine this data with CF and perform a random stratified split of 70:10:20 into train, validation and test sets to maintain similar class ratios. Finally, we merge this random test set with the held-out data for unseen task types, which creates our final test set of 364 data points. 

\subsection{Baselines}
We compare our approach to several baselines, as described in the following. With the first two baselines, we aim to evaluate how models pre-trained with other NLI datasets perform on the coherence classification problem. We design the last two baselines to understand the effect of our approach of modeling coherence assessment as an entailment recognition problem. To do so, we simply train a text-pair classifier, i.e., without performing transfer learning from other entailment recognition datasets. Our baselines are the following: 
\begin{itemize}
    \item RoBERTa-large-MNLI\footnote{https://huggingface.co/FacebookAI/roberta-large-mnli} - A language model based on the RoBERTa architecture, fine-tuned on MNLI~\cite{williams2018broad}.
    \item DeBERTa-v3-base-NLI\footnote{https://huggingface.co/MoritzLaurer/DeBERTa-v3-base-mnli-fever-anli} - A language model based on the DeBERTa-v3-base architecture that is fine-tuned on 763913 premise-hypothesis pairs from 3 NLI datasets.
    \item DeBERTa-v3-base - We obtain text-pair representation using DeBERTa-v3-base~\cite{he2022debertav3} and pass it to a randomly initialized dense layer to perform classification.
    \item BERT -  Similar to DeBERTa-v3-base, but we use~\cite{devlin-etal-2019-bert} to obtain text-pair representation.
\end{itemize}
Our model is based on DeBERTa-v3-base-NLI, but we train it with a coherence classification objective using our dataset. We fine-tune the last two baselines using the same training configuration as our model. We train for 3 epochs using a learning rate of $5e^{-5}$, batch size of 8, weight decay = 0.02, label smoothing $\alpha=0.05$, and using Adam optimizer. 

\subsection{Results}
\begin{table}[]
    \centering
    \begin{tabular}{|l|c|c|c|c|}
    \hline
    Model  & $\mathcal{E}^{g} \bot \mathcal{E}^t$  &$\mathcal{E}^g \models \mathcal{E}^t$  &$\mathcal{E}^g \not\models \mathcal{E}^t$  & F1\textsuperscript{macro} \\ 
    \hline \hline
    RoBERTa-large-MNLI   &0.26     &0.11   &0.49    &0.29 \\
    DeBERTa-v3-base-NLI     &0.24    &0.03 &0.42 &0.23 \\
    \hline
    DeBERTa-v3-base     &0.58    &0.58   &0.84   &0.67 \\
    BERT    &0.58   &0.56   &0.81   &0.65 \\
    \hline
    \textbf{Ours}      &0.87 &0.85 &0.91 &0.87 \\
    \hline
    \end{tabular}
    \caption{F1 scores for coherence classification. The first two sections denote only NLI training and only coherence classification training. Our model is first trained for NLI and then for coherence classification.}
    \label{tab:main-result}
\end{table}

\begin{table}[]
    \centering
    \begin{tabular}{|l|c|c|c|c|}
    \hline
    Model &  $\mathcal{E}^\pi \bot \mathcal{E}^t $   &$\mathcal{E}^\pi \models \mathcal{E}^t $    &$\mathcal{E}^\pi \not \models \mathcal{E}^t $ & F1\textsuperscript{macro} \\ 
    \hline \hline
    DeBERTa-v3-base     &0.38    &0.42   &0.91   &0.57 \\
    BERT    &0.42   &0.45   &0.85   &0.57 \\
    \hline
    \textbf{Ours}        &0.81   &0.84   &0.94   &0.86 \\
    \hline
    \end{tabular}
    \caption{Classification performance on $\mathcal{E}^\pi, \mathcal{E}^t$ pairs.}
    \label{tab:plan-result}
\end{table}

\begin{table}[t]
    \centering
    \begin{tabular}{|l|c|c|c|c|}
    \hline
    Model & $\mathcal{E}^{O_i} \bot \mathcal{E}^t $   &$\mathcal{E}^{O_i} \models \mathcal{E}^t $    &$\mathcal{E}^{O_i} \not \models \mathcal{E}^t $   & F1\textsuperscript{macro} \\ 
    \hline \hline
    DeBERTa-v3-base     &0.68    &0.66   &0.67   &0.67 \\
    BERT    &0.67   &0.63   &0.71   &0.67 \\
    \hline
    \textbf{Ours}        &0.90   &0.85   &0.84   &0.86 \\
    \hline
    \end{tabular}
    \caption{Classification performance on $\mathcal{E}^{O_i}, \mathcal{E}^t$ pairs.}
    \label{tab:obs-result}
\end{table}
We select the checkpoint having the highest macro-F1 score on the validation set and evaluate it on the test set. Table~\ref{tab:main-result} summarizes the main results. We find that models that are trained only on NLI datasets do not perform well for coherence classification. This is not unexpected because even though the two problems are similar, the existing NLI datasets contain data from domains that are unrelated to robotics. Also, our definition of entailment recognition differs from the definition in existing NLI annotation schemes. We further find that both DeBERTa-v3-base and BERT perform much better than pre-trained NLI models when trained on our dataset. Both models perform similarly, but the fine-tuned DeBERTa-v3-base has slightly better scores on $\mathcal{E}^g \models \mathcal{E}^t$ and $\mathcal{E}^g \not\models \mathcal{E}^t$. Finally, we find that our approach of fine-tuning, after pre-training to perform NLI, works well for the coherence classification problem. The results also support our decision to model coherence classification as an entailment recognition problem, as we find that the representations learned by training on NLI datasets help to improve coherence classification accuracy.

We further analyze the results separately for $\mathcal{E}^\pi$ and $\mathcal{E}^{O_i}$, as they require reasoning on different types of information (sequential vs. non-sequential). As shown in Table~\ref{tab:plan-result} and Table~\ref{tab:obs-result}, the models generally perform better in classifying $\mathcal{E}^{O_i},\mathcal{E}^t$ pairs. We believe this is because the models have to perform reasoning on sequential information (i.e., the plan) for $\mathcal{E}^\pi$, which is more difficult than $\mathcal{E}^{O_i}$ which is only a set of observations. Nevertheless, these results further support the efficacy of our approach as our models perform much better than the baselines, particularly for $\bot$ and $\models$ classes, which are more important than the $\not \models$ classes. Finally, we show the evaluation results on $\mathcal{E}^g, \mathcal{E}^t$ pairs from the held-out subset of new task types in Table~\ref{tab:unseen-result}. The results indicate that our model shows a better capability of generalization to explanations from unseen task types, outperforming both baselines by a large margin.

\begin{table}[]
    \centering
    \begin{tabular}{|l|c|c|c|c|}
    \hline
    Model   &$\mathcal{E}^{g} \bot \mathcal{E}^t$  &$\mathcal{E}^g \models \mathcal{E}^t$  &$\mathcal{E}^g \not\models \mathcal{E}^t$ & F1\textsuperscript{macro} \\ 
    \hline \hline
    DeBERTa-v3-base     &0.32    &0.36   &0.49   &0.39 \\
    BERT    &0.24   &0.35   &0.29   &0.30 \\
    \hline
    \textbf{Ours}        &0.52   &0.61   &0.66   &0.59 \\
    \hline
    \end{tabular}
    \caption{Coherence classification on $\mathcal{E}^g, \mathcal{E}^t$ pairs from tasks that are unseen during training.}
    \label{tab:unseen-result}
\end{table}
\subsection{Future Work}
\label{sec:discuss}
In this work, we have only discussed explanations of robot failures, but the problem of multimodal coherence can be studied beyond failures and explanations, e.g., multimodal communication in HRI. Second, coherence assessment being the focus of this work, we have proposed simple strategies for refining incoherent explanations. Future work can explore more complex strategies, such as re-prompting the LLM with the source of contradiction and dialog-based refinement. Third, we have defined the coherence taxonomy using a simple conjunction of propositions. However, explanations may contain dis-junctions and other complex logical structures which should be studied as well. Finally, we plan to perform user studies to understand the effect of incoherent multimodal explanations and their refinements using subjective measures.

\section{Conclusion}
In this work, we introduce and formulate the problem of detecting coherence in multimodal explanations of robot failures. We observe that a simple combination of explanations from multiple modalities is not sufficient to produce a coherent explanation. We propose an approach to detect if a pair of explanations is coherent and apply this method to a multimodal explanation generation framework that provides explanations by combining natural language, scene graph, and sequence of actions executed by a robot. In particular, we model coherence assessment as a logical entailment recognition problem and propose to solve it as a classification problem. Our experiments suggest that this modeling is beneficial, as we find that fine-tuning a model that was previously trained to detect textual entailment in other domains is an efficient approach to training an accurate coherence classifier. Further, we propose refinement strategies to convert incoherent explanations to coherent ones. 

\balance
\bibliographystyle{IEEEtran}
\bibliography{references}

\end{document}